\documentclass[12pt]{article}
\usepackage{amsfonts}
\usepackage{openwork}
\usepackage{booktabs}
\usepackage{longtable}
\usepackage[most]{tcolorbox}
\usepackage{adjustbox}
\usepackage{ragged2e}

\makeatletter
\renewcommand{\fnum@table}{\small\textbf{\tablename~\thetable}}
\renewcommand{\fnum@figure}{\small\textbf{\figurename~\thefigure}}
\newcommand{\fixedfirstpagefootnote}[1]{%
  \begingroup
  \renewcommand\thefootnote{}%
  \renewcommand\@makefnmark{}%
  \long\def\@makefntext##1{\noindent ##1}%
  \footnotetext{#1}%
  \addtocounter{footnote}{-1}%
  \endgroup
}
\makeatother

\title{AI's Blind Spots: Geographic Knowledge and Diversity Deficit in Generated Urban Scenario}
\author[*,**]{Ciro Beneduce}
\author[*]{Massimiliano Luca}
\author[*]{Bruno Lepri}
\affil[*]{ Bruno Kessler Foundation}
\affil[**]{University of Trento}
\affil[]{\texttt{cbeneduce@fbk.eu}, \texttt{mluca@fbk.eu}, \texttt{lepri@fbk.eu}}
\date{}

\begin{document}

\maketitle
\fixedfirstpagefootnote{Published in ``Proceedings of the 1st International Conference on Geospatial Artificial Intelligence (GeoAI 2026) -- Oral Presentation Papers'', edited by Haosheng Huang and Nico Van de Weghe, GeoAI 2026, 3--6 June 2026, Ghent, Belgium. \newline\newline
This contribution underwent single-blind peer review based on the extended abstract.
}

\begin{abstract}
Diffusion-based text-to-image models are increasingly used for urban analysis and scenario generation, but their geographic knowledge and representational biases remain poorly understood. We evaluate FLUX 1-\textit{schnell} and Stable Diffusion 3.5-Large in the United States by generating 150 street-view images for each state, each state capital, and a generic ``USA'' prompt. Images are embedded with DINO-v2 ViT-S/14 and compared with Fr\'echet Inception Distance (FID). Pairwise FID clustering shows that geographically proximate states and capitals often group together, indicating implicit geographic structure. However, the generic ``USA'' prompt collapses this diversity into a metropolitan stereotype: frontier, desert, tropical, rural, and small-city environments are underrepresented or distant in FID space. These results show that diffusion models can encode fine-grained geography while still reproducing narrow national-scale visual stereotypes.\\

\end{abstract}

\section{Introduction}\label{sec:intro}

Generative artificial intelligence is becoming part of the computational and visual infrastructure through which urban spaces are analyzed, simulated, and imagined. In urban spatial analysis, generative models are increasingly used for scenario generation, movement-related applications, and planning workflows \cite{wang2025generativeaiurbanplanning,yang2024uncoveringhumanmotionpattern,beneduce2024largelanguagemodelszeroshot,hall2024geographicinclusionevaluationtexttoimage,beneduce2025urban}. In the visual domain, diffusion models can produce realistic and context-rich imagery \cite{rombach2022high}, supporting applications such as image editing \cite{meng2022sdeditguidedimagesynthesis}, artistic and photorealistic generation \cite{nichol2022glidephotorealisticimagegeneration}, and text-to-video synthesis \cite{ho2020denoisingdiffusionprobabilisticmodels,singer2022makeavideotexttovideogenerationtextvideo}. As these systems become more widely used to represent places, their outputs may influence how urban environments are communicated, compared, and imagined \cite{hall2024geographicinclusionevaluationtexttoimage, wan2024surveybiastexttoimagegeneration}. This raises a central question for urban and geospatial applications: not only whether text-to-image models can produce plausible urban scenes, but also which geographies they represent well, which they simplify, and which they leave at the margins.

This question belongs to a broader debate on bias in generative AI. Bias has been widely studied in language models, whose outputs can reproduce stereotypes, unequal associations, and socially patterned inferences across dimensions such as gender, race, religion, nationality, occupation, and everyday behavior \cite{nadeem2021stereoset,parrish2022bbq,luca2025llmwearspradaanalysing,smith2022m,gallegos2024bias}. Similar concerns apply to diffusion models, where biased associations may be encoded visually rather than textually \cite{luccioni2023stable}. In urban and geospatial settings, this visual dimension is especially consequential because generated images are not neutral depictions.
Geographic bias therefore concerns not only whether a generated image is realistic, but also whether it captures the diversity of the place it claims to represent.  If a country is repeatedly represented through dense downtown streets, metropolitan skylines, and glass-fronted buildings, non-metropolitan landscapes such as rural settlements, smaller cities, desert regions, tropical environments, and frontier geographies may be displaced by a narrow national stereotype. 

Recent work shows that this concern is not merely hypothetical. Vision and text-to-image systems can reinforce stereotypes or misrepresent groups and regions, partly because their training data reflect uneven visibility across people, cultures, and places \cite{jha2024visageglobalscaleanalysisvisual,wan2024surveybiastexttoimagegeneration}. In spatial settings, evaluations have found regional disparities in text-to-image outputs, with some areas, such as Africa or West Asia, receiving less realistic or less diverse visual representations than others \cite{hall2024diginevaluatingdisparities}. These studies establish geographic bias as an important dimension of generative model evaluation. What remains less clear is whether diffusion models preserve geographic variation across spatial scales.

A key issue is the distinction between geographic knowledge and geographic diversity. A model may encode meaningful associations between place names and visual features, distinguishing, for example, between desert states, coastal cities, mountain regions, and large metropolitan capitals when these places are named explicitly. Yet this does not imply that a generic prompt will preserve the same internal variation. The model may instead collapse national representation into a narrower visual prototype. We define this as a scale-dependent representational failure: geographic distinctions are available at specific spatial scales, but are not carried into broader geographic categories.

In this paper, we examine this scale-dependent tension through a systematic analysis of generated U.S. street-view imagery. The United States provides a useful test case because it contains substantial internal geographic and urban diversity, including dense metropolitan regions, small capitals, rural settlements, tropical and desert landscapes, mountain states, frontier environments, and coastal cities. We analyze two open text-to-image diffusion models: FLUX 1-\textit{schnell} \cite{flux2024} and Stable Diffusion 3.5-\textit{Large} \cite{rombach2022high}. For each model, we generate 150 images for every U.S. state, every state capital, and the generic prompt ``USA'', yielding 15,150 images per model. We encode the images with DINO-v2 ViT-S/14 \cite{oquab2024dinov2learningrobustvisual} and measure location-to-location similarity using Fr\'echet Inception Distance (FID) \cite{heusel2018ganstrainedtimescaleupdate}. This design allows us to test whether generated imagery contains coherent geographic structure and whether the generic national prompt preserves or collapses the visual diversity of specific places.

Our results reveal a dual pattern. When prompted with specific states and capitals, both models generate images whose embedding structure often reflects geographic proximity and shared built-environment characteristics, suggesting implicit geographic knowledge. By contrast, when prompted with ``USA'', both models produce a narrower, metropolis-like representation. Rural areas, smaller cities, frontier landscapes, desert regions, and tropical environments remain visually distant from the generic national prompt, even though the same models can differentiate many of these places when prompted directly. These findings show that diffusion models can encode fine-grained geographic distinctions while still producing narrow national-scale stereotypes. More broadly, they suggest that evaluating generative AI for urban and geospatial applications requires moving beyond realism and prompt adherence toward measures of geographic diversity, spatial coverage, and representational inclusion.

\begin{figure}
\centering
\includegraphics[width=1\linewidth]{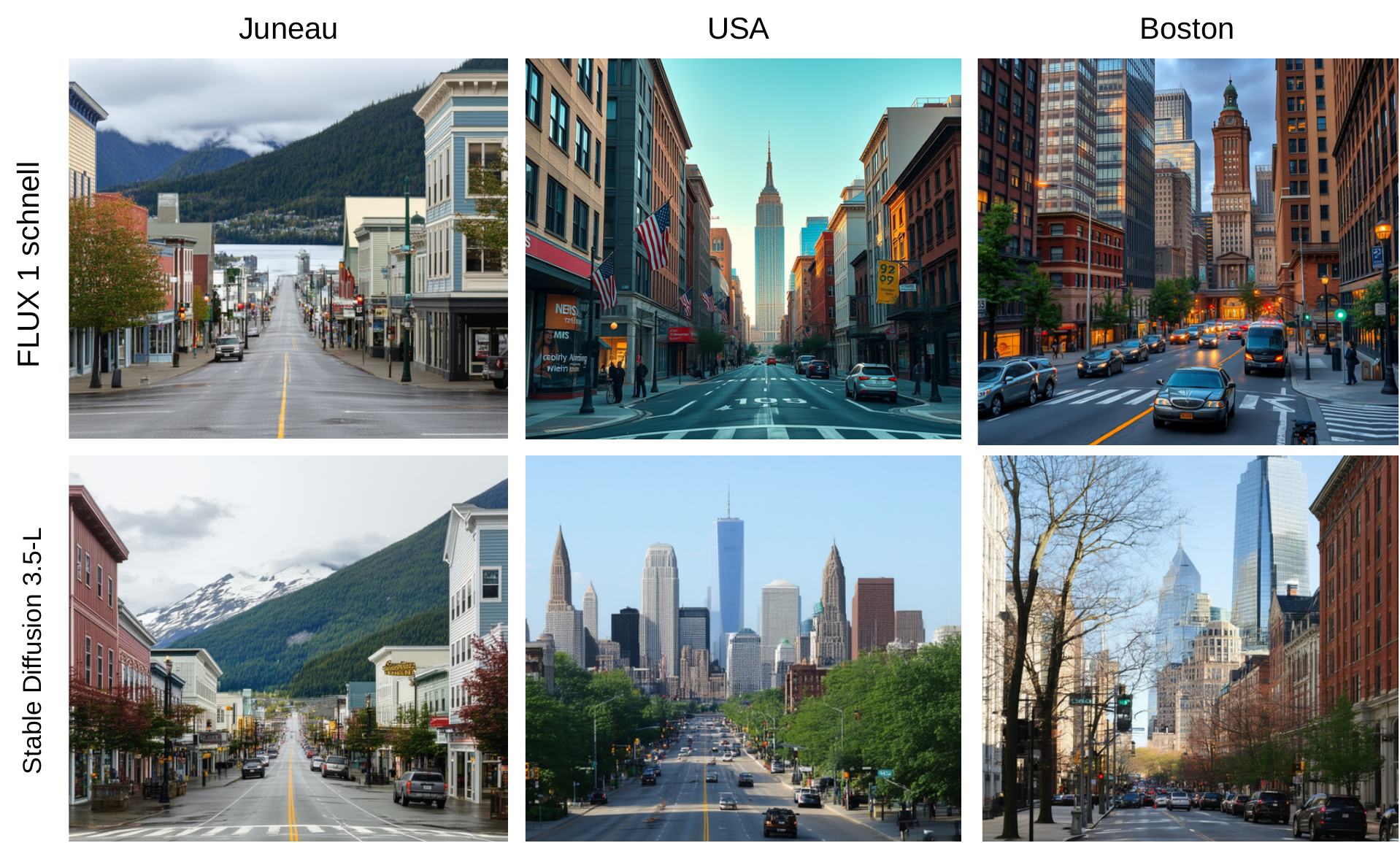}
\caption{Outputs from FLUX 1-schnell (top) and Stable Diffusion 3.5-L (bottom) for three prompts: ``USA,'' ``Boston,'' and ``Juneau''. Cities are generated plausibly, while ``USA'' defaults to a stereotypical metropolitan scenario.}
\label{fig:network_keywords}
\end{figure}

\section{Methodology}\label{sec:methods}

\subsubsection{Models and Prompt}\label{sec:models}
To analyze place identity in synthetic imagery, we employ two representative open-weights diffusion transformers: FLUX 1-\textit{schnell} and Stable Diffusion 3.5-L. FLUX 1-\textit{schnell}~\cite{flux2024} is a 12B-parameter rectified flow transformer optimized via latent adversarial diffusion distillation, enabling high-fidelity synthesis in few sampling steps. Stable Diffusion 3.5-L, an 8.1B-parameter multimodal diffusion transformer, is selected for its enhanced prompt adherence and photorealism capabilities.

We use a single standardized prompt template across all generations:

\begin{tcolorbox}[
    colback=gray!4,
    colframe=gray!55,
    colbacktitle=gray!18,
    coltitle=black,
    title={Prompt template},
    fonttitle=\bfseries\small,
    arc=2pt,
    boxrule=0.4pt,
    boxsep=2pt,
    left=2mm,
    right=2mm,
    top=1mm,
    bottom=1mm
]
\small
\texttt{A photorealistic high-resolution street-view photo of \{LOCATION\}}
\end{tcolorbox}

\noindent Here, \texttt{\{LOCATION\}} is instantiated with each of the 50 U.S. states, their respective capitals, and the generic token \texttt{"USA"}, yielding 101 distinct prompts. We generate 150 samples per prompt, resulting in 15,150 images per model.

The phrase ``street-view'' is chosen to encourage grounded, built-environment imagery, such as roads, buildings, vegetation, vehicles, signage, and urban or rural layouts. We intentionally omit additional stylistic modifiers, such as weather, lighting, season, demographic descriptors, or artistic style, in order to reduce prompt-induced confounds and better isolate the model's internal association between a location name and its expected visual appearance. This setup treats each generated image as a sample from the model's learned visual distribution for a given place name.

Inference parameters, including guidance scale and diffusion steps, are kept at model-specific defaults. The final generation settings are reported in Appendix Table ~\ref{tab:generation_settings}.

\subsubsection{Embedding}\label{sec:embedding}

To map the generated imagery into a semantically meaningful visual space, we use DINO-v2 ViT-S/14 \parencite{oquab2024dinov2learningrobustvisual} as a frozen feature extractor. DINO-v2 is trained with self-supervision and provides visual representations that capture object content, scene layout, texture, vegetation, road structure, and built-environment cues without relying on explicit geographic labels. This makes it suitable for our setting, where the goal is not to classify places directly, but to compare the visual distributions that models associate with different place names.

Each generated image is resized to 256 pixels on the shortest side, centre-cropped to 224$\times$224, converted to a tensor, and normalised using the standard mean and variance. We then pass the image through the DINO-v2 encoder and extract one image-level feature vector. 

Let $z_i^{(\ell)} \in \mathbb{R}^d$ denote the embedding of image $i$ generated for location $\ell$. For each location, we summarise the set of 150 embeddings by its empirical mean vector $\mu_\ell$ and covariance matrix $\Sigma_\ell$:
\[
\mu_\ell = \frac{1}{N}\sum_{i=1}^{N} z_i^{(\ell)}, 
\qquad
\Sigma_\ell = \frac{1}{N}\sum_{i=1}^{N}
\left(z_i^{(\ell)}-\mu_\ell\right)
\left(z_i^{(\ell)}-\mu_\ell\right)^\top ,
\]
where $N=150$.

The mean vector captures the central visual tendency of a location, while the covariance matrix captures the variability of generated scenes around that tendency. This distributional representation is important because a location prompt may produce a heterogeneous set of images rather than a single visual prototype. For example, two locations may have similar average streetscapes but differ in the diversity of generated architecture, vegetation, road layout, or urban density. Representing each location by both mean and covariance therefore allows us to compare not only central similarity, but also the spread of the model's visual associations.

\subsubsection{Distance Metrics}\label{sec:distance}

To compare locations, we compute a Fréchet distance between the Gaussian distributions defined by each location's DINO-v2 embedding mean and covariance. This follows the same distributional principle as the Fréchet Inception Distance (FID) \parencite{heusel2018ganstrainedtimescaleupdate}, but is computed in DINO-v2 feature space. For two locations with embedding summaries $(\mu_1,\Sigma_1)$ and $(\mu_2,\Sigma_2)$, we compute:
\begin{equation}
d(\mu_1, \Sigma_1; \mu_2, \Sigma_2) =
\|\mu_1 - \mu_2\|_2^2 + \mathrm{Tr}\left(\Sigma_1 + \Sigma_2 - 2\left(\Sigma_1 \Sigma_2\right)^{1/2}\right).
\end{equation}

Lower values indicate greater similarity between two generated visual distributions. The first term measures the distance between the average visual representations of two locations, while the covariance term accounts for differences in within-location variability. This is important for our setting because two locations may have similar mean embeddings while differing in the diversity of generated streetscapes, architectural forms, vegetation, or road layouts. Conversely, a location with a narrow and stereotyped set of generations may be distinguishable not only by its average representation, but also by its reduced dispersion in feature space.

We compute pairwise distances among all state prompts and among all capital prompts, yielding model-specific distance matrices for each geographic level. These matrices are used for hierarchical clustering to assess whether generated visual distributions recover coherent geographic or built-environment groupings. We also compute the distance from each state and capital to the generic \texttt{USA} prompt. This second comparison measures how closely each specific place aligns with the model's national-scale visual prototype, allowing us to identify which geographies are included in, or excluded from, the generic representation of the United States.

\section{Results}\label{sec:results}

We organize the results around two questions. First, do the generated images preserve meaningful geographic structure when locations are named explicitly? Second, does the generic \texttt{USA} prompt preserve this diversity, or does it collapse toward a narrower national prototype? We answer the first question using pairwise FID matrices and hierarchical clustering, and the second by measuring each state and capital's distance to the generic \texttt{USA} prompt.

\subsubsection{Evidence of Geographic Structure}
\begin{figure}[htbp]
    \centering
    \IfFileExists{C.pdf}{%
        \includegraphics[width=1\linewidth]{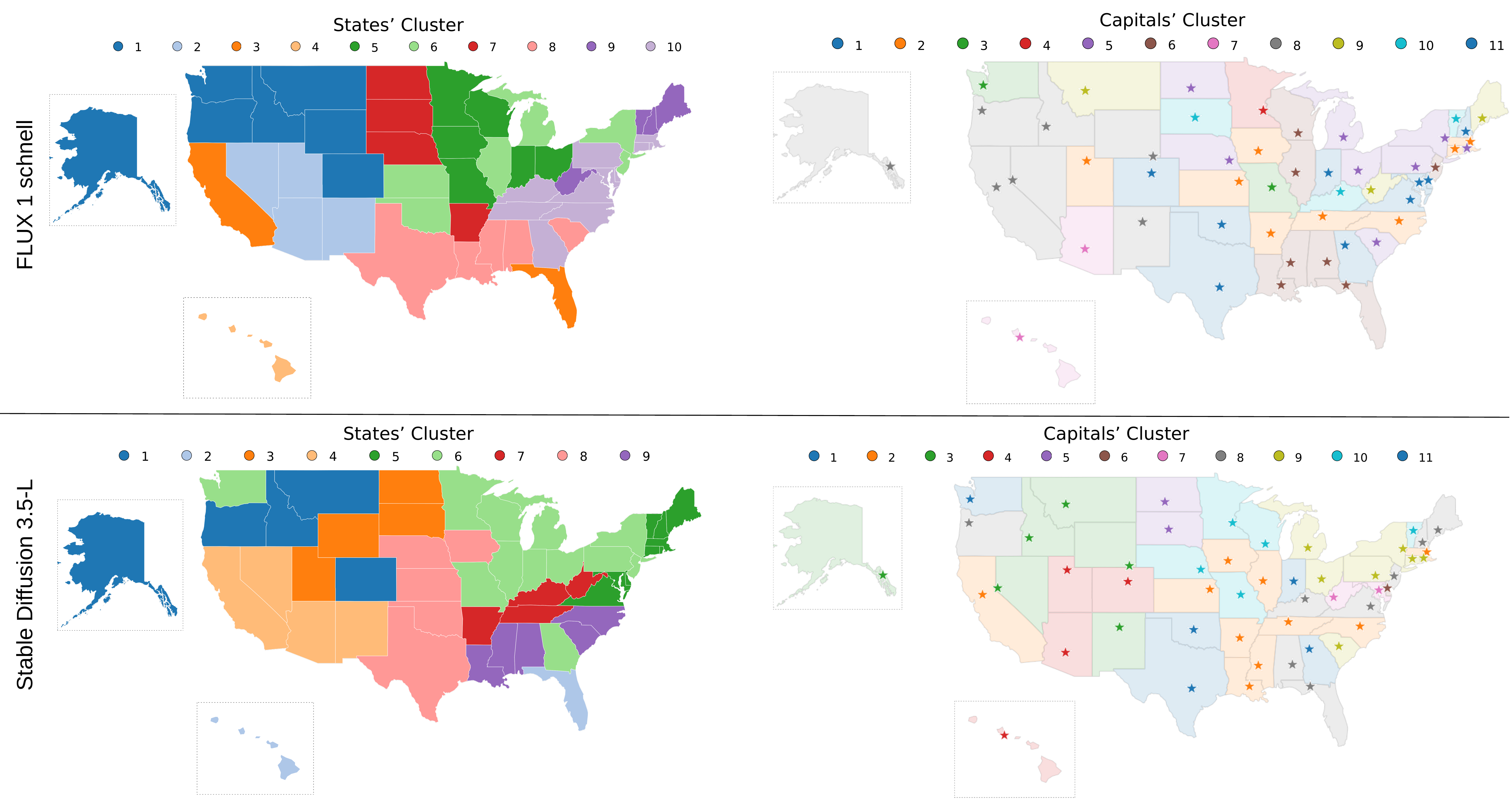}%
    }{%
        \fbox{\begin{minipage}[c][0.3\textheight][c]{0.9\linewidth}
        \centering \texttt{C.pdf} not available in the workspace.
        \end{minipage}}%
    }
    \caption{Maps of clusters derived from hierarchical clustering of FID matrices. The upper maps show FLUX 1-\textit{schnell}, while bottom ones refers to Stable Diffusion 3.5-L, both  with state-prompt clusters on the left and capital-prompt clusters on the right. In the state maps, each state polygon is colored according to its assigned cluster. In the capital maps, each marker is colored according to the cluster assigned to the corresponding state capital, while state boundaries are shown only for geographic reference. Cluster numbers are specific to each map.}
    \label{fig:Clusters}
\end{figure}

Pairwise FID show that both models encode meaningful geographic and built-environment relationships. As Figure~\ref{fig:Clusters} illustrates, states that share regional, climatic, or urban-form characteristics often appear in the same cluster. This suggests that the models do not treat U.S. place names as arbitrary labels. Instead, the generated image distributions contain a latent spatial structure that partially reflects real geographic variation.

For FLUX 1-\textit{schnell}, a Mountain West and frontier-oriented cluster includes Alaska, Colorado, Idaho, Montana, Oregon, Washington, and Wyoming. These states share visual cues such as mountainous terrain, lower-density development, coniferous vegetation, and open-road landscapes. A distinct Desert Southwest cluster contains Arizona, Nevada, New Mexico, and Utah, consistent with arid landscapes, sparse vegetation, and characteristic road and built-form patterns. Stable Diffusion 3.5-L shows a similar but not identical organization: it separates a core Mountain West group from a Plains/Mountain border grouping and places Arizona, California, Nevada, and New Mexico in a broader southwestern cluster. This indicates that both models encode geographic structure, but they partition transitional regions differently.

New England provides another example. FLUX 1-\textit{schnell} forms a compact grouping of Maine, New Hampshire, and Vermont, while Stable Diffusion 3.5-L groups parts of New England with Mid-Atlantic states. These differences suggest that the models' representations are shaped not only by physical proximity, but also by recurring visual templates such as small-town streets, older residential architecture, forested environments, and dense northeastern urban form.

Capital-city clustering similarly reflects urban hierarchy. Both models group several large or highly urbanized capitals, including Atlanta, Austin, Indianapolis, and Oklahoma City. Stable Diffusion 3.5-L further forms a western metropolitan capital cluster including Denver, Honolulu, Phoenix, and Salt Lake City. Mid-sized capitals also cluster coherently, suggesting that the models capture regularities in downtown scale, street width, commercial density, and civic architecture. Overall, the clustering results indicate that the models possess a degree of place-specific geographic knowledge. However, this knowledge is uneven and becomes fragile for smaller or more ambiguous capitals, as discussed next.

\subsubsection{Small Capital Misgeneration}

The clustering structure also reveals systematic failure modes for smaller capitals. Several capitals are not merely distant from otherwise similar U.S. cities; they are grouped together because they share incorrect or ambiguous visual cues. In FLUX 1-\textit{schnell}, three of four misgenerated capitals (Frankfort, Montpelier, Pierre) cluster together (Cluster 10), consistent with shared European architectural cues; Dover appears elsewhere (Cluster 11), alongside correctly generated U.S. capitals (Annapolis, Concord, Richmond), suggesting partial overlap in scale and streetscape characteristics. SD-3.5-L shows a structured pattern: Bismarck and Pierre cluster together (Cluster 5), Dover and Olympia appear as single-member clusters (Clusters 6 and 11), and Olympia is notably characterized by ancient Greek-like architecture. 

This failure mode is important because it differs from ordinary noise. The affected capitals are not randomly distributed across the embedding space; rather, they form structured outliers. This suggests that misgeneration can itself be systematic, producing coherent but incorrect visual prototypes for underrepresented or ambiguous places.

\subsubsection{Lack of Diversity}
Bias becomes most visible when comparing each state and capital to the generic ``USA'' prompt. Using FID to ``USA,'' both models represent ``USA'' primarily as dense urban environments, aligning closely with metropolitan states and cities while diverging strongly for frontier, desert, and tropical landscapes. \textit{Table~\ref{tab:fid_bias_unified}} shows that Alaska, Hawaii, Arizona, and Nevada are among the farthest from ``USA'' in FID space, with scores up to eight times higher than the metropolitan tier. The pattern extends to capitals: larger capitals, such as Raleigh and Boston, align closely with ``USA,'' while smaller capitals, such as Juneau, diverge substantially. Thus, despite encoding fine-grained geographic distinctions, the models deploy a narrow, urban-centric stereotype when prompted at the national scale.

%\subsubsection{The Generic \texttt{USA} as a Metropolitan Prototype}
The strongest evidence of geographic bias emerges when specific places are compared with the generic \texttt{USA} prompt. If the national prompt preserved the diversity encoded at the state and capital levels, we would expect a broad range of U.S. geographies to appear relatively close to \texttt{USA}. Instead, both models position \texttt{USA} near a narrow subset of locations associated with metropolitan or highly urbanised imagery.

Table~\ref{tab:fid_bias_unified} reports the five closest and five farthest states and capitals for each model; complete rankings are provided in Appendix~\ref{app:usa_distance_rankings}. For FLUX 1-\textit{schnell}, the closest state prompts include New Jersey, Michigan, Illinois, Minnesota, and Texas, while the closest capitals include Madison, Raleigh, Jackson, Nashville, and Little Rock. For Stable Diffusion 3.5-L, the closest states are Illinois, Minnesota, New York, Georgia, and Indiana, and the closest capitals are Indianapolis, Austin, Boston, Raleigh, and Sacramento. Although these locations are not identical across models, they share a tendency toward urban, suburban, or metropolitan streetscape imagery.

By contrast, frontier, tropical, desert, and visually distinctive regions are consistently distant from the generic national prototype. Hawaii and Alaska are among the farthest states for FLUX 1-\textit{schnell}, while Arizona, North Dakota, Hawaii, Nevada, and Wyoming are farthest for Stable Diffusion 3.5-L. Similar patterns appear at the capital level: Juneau, Pierre, Dover, Bismarck, Olympia, Montpelier, and Frankfort are among the most distant capitals across the two models. These locations correspond to smaller cities, remote capitals, frontier geographies, or prompts affected by the misgeneration patterns described above.

The magnitude of the ranking differences is substantial. For FLUX 1-\textit{schnell}, the closest state to \texttt{USA} is New Jersey with a distance of 331.8, while Hawaii reaches 2594.5. For Stable Diffusion 3.5-L, Illinois is closest at 413.7, while Arizona reaches 3352.9. This long tail indicates that the generic national prompt is not an inclusive average over U.S. geographic diversity. Rather, it functions as a narrow national prototype that aligns with some urbanized places while excluding others.

These results reveal a scale-dependent representational failure. When prompted with specific states and capitals, the models often recover meaningful geographic distinctions. Yet when prompted at the national scale, this diversity is not preserved. The national prompt collapses toward a metropolitan visual stereotype, leaving frontier, desert, tropical, rural, and small-capital environments at the margins of the generated representation.

\begin{table}[htbp]
  \centering
  \small
  \caption{Top--5 and bottom--5 FID to the generic USA prompt, combining \textbf{states} and \textbf{capital cities} across both models.}
  \label{tab:fid_bias_unified}
  \setlength{\tabcolsep}{4pt}
  
  \begin{tabular}{@{} r l r l r @{\hspace{2.5em}} l r l r @{}}
    \toprule
             & \multicolumn{4}{c}{\textbf{FLUX 1-\textit{schnell}}}
             & \multicolumn{4}{c}{\textbf{Stable Diffusion 3.5-L}} \\
    \cmidrule(r){2-5} \cmidrule(l){6-9}
    \textbf{Rank}
      & \multicolumn{1}{c}{\textbf{State}} & \multicolumn{1}{c}{\textbf{FID}} & \multicolumn{1}{c}{\textbf{Capital}} & \multicolumn{1}{c}{\textbf{FID}}
      & \multicolumn{1}{c}{\textbf{State}} & \multicolumn{1}{c}{\textbf{FID}} & \multicolumn{1}{c}{\textbf{Capital}} & \multicolumn{1}{c}{\textbf{FID}} \\
    \midrule
    \multicolumn{9}{c}{\textit{Most similar to ``USA'' (lowest FID)}}\\
     1 & New Jersey     & 331.8 & Madison       & 408.5 & Illinois       & 413.7 & Indianapolis & 570.9 \\
     2 & Michigan       & 404.2 & Raleigh       & 467.5 & Minnesota      & 796.2 & Austin       & 645.2 \\
     3 & Illinois       & 414.1 & Jackson       & 484.7 & New York       & 829.2 & Boston       & 662.8 \\
     4 & Minnesota      & 545.8 & Nashville     & 501.1 & Georgia        & 843.5 & Raleigh      & 665.9 \\
     5 & Texas          & 599.4 & Little Rock   & 539.5 & Indiana        & 848.3 & Sacramento   & 705.3 \\
    \addlinespace
    \multicolumn{9}{c}{\textit{Least similar to ``USA'' (highest FID)}}\\
    50 & Hawaii         & 2594.5 & Frankfort     & 2968.3 & Arizona        & 3352.9 & Dover        & 3839.6 \\
    49 & Alaska         & 2143.4 & Saint Paul    & 2730.7 & North Dakota   & 3207.3 & Olympia      & 3430.7 \\
    48 & Florida        & 1988.7 & Dover         & 2530.8 & Hawaii         & 3139.8 & Bismarck     & 3192.8 \\
    47 & Arizona        & 1970.6 & Pierre        & 2522.3 & Nevada         & 3131.7 & Pierre       & 2962.8 \\
    46 & California     & 1961.9 & Montpelier    & 2481.1 & Wyoming        & 3088.3 & Juneau       & 2607.0 \\
    \bottomrule
  \end{tabular}
\end{table}

\section{Discussion and Conclusion}\label{sec:discussion}

Our findings reveal a scale-dependent tension in the geographic behavior of text-to-image models. When prompted with specific places, both FLUX 1-\textit{schnell} and Stable Diffusion 3.5-L generate imagery whose embedding structure often reflects regional proximity, shared landscapes, and similarities in built form. This suggests that these models encode a latent form of geographic knowledge: place names are not treated as arbitrary tokens, but are associated with recurring visual patterns such as desert environments, mountainous regions, dense downtowns, smaller civic centers, and northeastern urban form. In this sense, generated images can contain meaningful spatial structure.

At the same time, this geographic knowledge is not preserved when the prompt is generalized to the national scale. The generic \texttt{USA} prompt collapses toward a narrow metropolitan prototype, aligning most closely with urbanized states and capitals while remaining distant from frontier, tropical, desert, rural, and small-capital environments. This is the central representational failure identified in this paper: the models can distinguish many places when they are named explicitly, yet they do not aggregate this diversity into an inclusive national representation. Geographic bias therefore appears not only as incorrect depiction of individual places, but also as a failure to preserve diversity across spatial scales.

The misgeneration of small capitals further illustrates how this failure can become systematic. Places such as Pierre, Frankfort, Montpelier, Dover, Bismarck, and Olympia are not merely noisy cases. Their generated distributions often move toward coherent but incorrect visual templates, including European or otherwise globally salient architectural cues. This suggests that sparse visual evidence, ambiguous toponyms, or uneven representation in web-scale training data can push models toward more dominant associations. 

Our study also suggests that evaluating text-to-image models only through realism or prompt adherence is insufficient. A generated image of \texttt{USA} may be photorealistic and apparently plausible while still being geographically narrow. Conversely, a model may show strong place-specific knowledge while failing to represent broader geographic categories inclusively. Evaluations should therefore consider both specificity and aggregation: whether models can represent named places accurately, and whether broader prompts preserve the diversity of the places they subsume.

Several limitations remain. First, our analysis focuses on the United States. This setting provides substantial internal geographic diversity, but it does not capture the full global unevenness of image-text data. The representational failures observed here may be stronger for countries and regions that are less visible in web-scale datasets or are more often represented through stereotyped imagery. Second, our prompt template intentionally avoids stylistic modifiers in order to isolate the effect of location names. Real users may use richer prompts involving season, weather, safety, architecture, people, or planning scenarios, which could alter the resulting distributions. Finally, DINO-v2 captures semantic and layout-level similarity, but it does not directly measure geographic correctness. Human evaluation and comparisons with curated geotagged reference imagery would provide useful complementary validation.

Future work should extend this framework beyond the United States and evaluate geographic representation across countries, regions, and cities with different levels of visibility in training data. Another important direction is to test mitigation strategies, including prompt diversification, geographically balanced reference sets, retrieval-augmented generation, and post-generation auditing. More broadly, generative models used in urban and geospatial contexts should be evaluated not only for image quality, but also for geographic coverage, representational diversity, and inclusion.

In conclusion, our results show that open text-to-image diffusion models encode meaningful geographic distinctions while simultaneously producing narrow national-scale stereotypes. This dual behavior matters: the presence of place-specific knowledge does not guarantee inclusive geographic representation. For generative AI to be responsibly used in urban analysis and civic imagination, models must be assessed on which places they make visible, which they compress into stereotypes, and which they leave out.

\section{Acknowledgements}
B.L. and M.L. acknowledge the partial support of the European Union's Horizon Europe research and innovation program under grant agreement No. 101120237 (ELIAS), and  B.L. acknowledges the partial support of the European Union's Horizon Europe research and innovation program under grant agreement No. 101120763 (TANGO).

\section{References}
\printbibliography[heading=none]

% Requires:
% \usepackage{booktabs}

\clearpage
\appendix
\section{Appendix}
\section{Generation settings}
\label{app:generation_settings}

Table~\ref{tab:generation_settings} reports the model-specific generation settings used for all synthetic image generation. The prompt template and prompt set are described in Section~\ref{sec:models}; here we report only the sampling parameters required for reproducibility.

\begin{table}[htbp]
\centering
\small
\setlength{\tabcolsep}{7pt}
\renewcommand{\arraystretch}{1.12}
\caption{Model-specific generation settings.}
\label{tab:generation_settings}
\begin{tabular}{lcc}
\toprule
\textbf{Setting} & \textbf{FLUX 1-\textit{schnell}} & \textbf{Stable Diffusion 3.5-L} \\
\midrule
Samples per prompt & 150 & 150 \\
Batch size & 5 & 5 \\
Inference steps & 4 & 28 \\
Guidance scale & 0.0 & 4.5 \\
Maximum sequence length & 256 & 512 \\
Precision & \texttt{bfloat16} & \texttt{bfloat16} \\
Seed schedule & \multicolumn{2}{c}{\texttt{42 + batch index}} \\
\bottomrule
\end{tabular}
\end{table}

\section{Distance matrix construction}
\label{app:matrix_construction}

For each model and geographic level, we construct a separate FID matrix. The state-level matrices contain the 50 U.S. state prompts together with the generic \texttt{USA} prompt. The capital-level matrices contain the 50 state-capital prompts. For each location $\ell$, the 150 generated images are embedded with DINO-v2 ViT-S/14 and summarized by a mean vector $\mu_\ell$ and covariance matrix $\Sigma_\ell$.

Given two locations $\ell_i$ and $\ell_j$, we compute their FID using the Gaussian summaries $(\mu_i,\Sigma_i)$ and $(\mu_j,\Sigma_j)$, as defined in Section~\ref{sec:distance}. Repeating this calculation for every pair of locations yields a symmetric matrix $D$, where $D_{ij}$ is the visual distributional distance between locations $\ell_i$ and $\ell_j$. Diagonal entries are zero by construction.

We construct four primary matrices:
\begin{itemize}
    \item FLUX 1-\textit{schnell}, state prompts;
    \item FLUX 1-\textit{schnell}, capital prompts;
    \item Stable Diffusion 3.5-L, state prompts;
    \item Stable Diffusion 3.5-L, capital prompts.
\end{itemize}

The state matrices include the generic \texttt{USA} prompt and are therefore used both for pairwise state comparison and for computing each state's distance to \texttt{USA}. Capital-to-\texttt{USA} distances are computed analogously by comparing each capital's DINO-v2 embedding distribution with the distribution generated from the generic \texttt{USA} prompt. Complete rankings are reported in Appendix~\ref{app:usa_distance_rankings}.

\section{Hierarchical clustering and map construction}
\label{app:clustering_procedure}

The cluster maps in Figure~\ref{fig:Clusters} are constructed from the FID matrices described in Appendix~\ref{app:matrix_construction}. Clustering is performed separately for each model and prompt level: FLUX.1-\textit{schnell} states, FLUX.1-\textit{schnell} capitals, SD-3.5-Large states, and SD-3.5-Large capitals.

For each distance matrix \(D\), we first apply an element-wise square-root transformation, \(\sqrt{D}\). This transformation reduces the dominance of very large pairwise distances while preserving the ordering of distances. The transformed matrix is then converted to condensed distance-vector form and clustered using Ward-linkage agglomerative clustering:
\[
Z = \mathrm{Ward}\left(\mathrm{squareform}(\sqrt{D})\right).
\]
The resulting dendrogram is cut at a distance threshold \(\tau\), producing the cluster labels used in the maps. For state-level clustering, the generic \texttt{USA} prompt is excluded before clustering so that the maps represent relationships among the 50 states only. Capital-level clustering is performed over the 50 capital prompts.

Table~\ref{tab:clustering_settings} reports the cut thresholds and resulting number of clusters used for the maps. Small changes within the same interval between dendrogram merge heights can produce identical cluster assignments; therefore, the reported thresholds should be interpreted as the effective cut values used to obtain the displayed partitions. Exact cluster memberships are reported in Appendix~\ref{app:cluster_composition}.

\begin{table}[ht]
\centering
\small
\setlength{\tabcolsep}{6pt}
\renewcommand{\arraystretch}{1.1}
\caption{Hierarchical clustering settings used for the cluster maps.}
\label{tab:clustering_settings}
\begin{tabular}{llrr}
\toprule
\textbf{Model} & \textbf{Prompt level} & \textbf{Cut threshold \(\tau\)} & \textbf{Clusters \(k\)} \\
\midrule
FLUX.1-\textit{schnell} & State   & 35.0 & 10 \\
FLUX.1-\textit{schnell} & Capital & 40.0 & 11 \\
SD-3.5-Large            & State   & 52.0 & 9  \\
SD-3.5-Large            & Capital & 42.0 & 11 \\
\bottomrule
\end{tabular}
\end{table}

The resulting clusters are used as descriptive summaries of the generated visual structure. They should not be interpreted as ground-truth geographic regions. Instead, they identify groups of locations whose generated image distributions are close in DINO-v2 feature space. Cluster labels are local to each model and prompt level: for example, Cluster 1 in the FLUX.1-\textit{schnell} state map is not directly comparable to Cluster 1 in the SD-3.5-Large capital map.

\section{Cluster composition}
\label{app:cluster_composition}

Tables~\ref{tab:cluster_composition_flux1_schnell_state}--\ref{tab:cluster_composition_sd_35_large_capital}
report the exact cluster memberships used to construct Figure~\ref{fig:Clusters}. Each table corresponds to one model and one prompt level. These tables provide the definitive record of the cluster labels used in the geographic visualizations.

%%%%%%%%%%%%%%%%%%%%%%%%%%%%%%%%%%%%%%%%%%%%%%%%%%%%%%%%%%%%%%%%%%%%%%%%%%%%%%%%%%%%%%%%%%
% FLUX.1-\textit{schnell} -- State
%%%%%%%%%%%%%%%%%%%%%%%%%%%%%%%%%%%%%%%%%%%%%%%%%%%%%%%%%%%%%%%%%%%%%%%%%%%%%%%%%%%%%%%%%%
\begin{table}[ht]
\centering
\scriptsize
\setlength{\tabcolsep}{4pt}
\renewcommand{\arraystretch}{0.95}
\caption{Cluster membership for state prompts in FLUX.1-\textit{schnell}.}
\label{tab:cluster_composition_flux1_schnell_state}
\begin{tabular}{r r p{0.68\linewidth}}
\toprule
\textbf{Cluster} & \textbf{Size} & \textbf{Locations} \\
\midrule
1 & 7 & Alaska, Colorado, Idaho, Montana, Oregon, State of Washington, Wyoming \\
2 & 4 & Arizona, Nevada, New Mexico, Utah \\
3 & 2 & California, Florida \\
4 & 1 & Hawaii \\
5 & 6 & Indiana, Iowa, Minnesota, Missouri, Ohio, Wisconsin \\
6 & 6 & Illinois, Kansas, Michigan, New Jersey, Oklahoma, State of New York \\
7 & 4 & Arkansas, Nebraska, North Dakota, South Dakota \\
8 & 5 & Alabama, Louisiana, Mississippi, South Carolina, Texas \\
9 & 4 & Maine, New Hampshire, Vermont, West Virginia \\
10 & 11 & Connecticut, Delaware, Georgia, Kentucky, Maryland, Massachusetts, North Carolina, Pennsylvania, Rhode Island, Tennessee, Virginia \\
\bottomrule
\end{tabular}
\end{table}

%%%%%%%%%%%%%%%%%%%%%%%%%%%%%%%%%%%%%%%%%%%%%%%%%%%%%%%%%%%%%%%%%%%%%%%%%%%%%%%%%%%%%%%%%%
% FLUX.1-\textit{schnell} -- Capital
%%%%%%%%%%%%%%%%%%%%%%%%%%%%%%%%%%%%%%%%%%%%%%%%%%%%%%%%%%%%%%%%%%%%%%%%%%%%%%%%%%%%%%%%%%
\begin{table}[ht]
\centering
\scriptsize
\setlength{\tabcolsep}{4pt}
\renewcommand{\arraystretch}{0.95}
\caption{Cluster membership for capital prompts in FLUX.1-\textit{schnell}.}
\label{tab:cluster_composition_flux1_schnell_capital}
\begin{tabular}{r r p{0.68\linewidth}}
\toprule
\textbf{Cluster} & \textbf{Size} & \textbf{Locations} \\
\midrule
1 & 5 & Atlanta, Austin, Denver, Indianapolis, Oklahoma City \\
2 & 8 & Boston, Des Moines, Hartford, Little Rock, Nashville, Raleigh, Salt Lake City, Topeka \\
3 & 2 & Jefferson City, Olympia \\
4 & 1 & Saint Paul \\
5 & 8 & Albany, Bismarck, Columbia, Columbus, Harrisburg, Lansing, Lincoln, Providence \\
6 & 7 & Baton Rouge, Jackson, Madison, Montgomery, Springfield, Tallahassee, Trenton \\
7 & 2 & Honolulu, Phoenix \\
8 & 7 & Boise, Carson City, Cheyenne, Juneau, Sacramento, Salem, Santa Fe \\
9 & 3 & Augusta, Charleston, Helena \\
10 & 3 & Frankfort, Montpelier, Pierre \\
11 & 4 & Annapolis, Concord, Dover, Richmond \\
\bottomrule
\end{tabular}
\end{table}

%%%%%%%%%%%%%%%%%%%%%%%%%%%%%%%%%%%%%%%%%%%%%%%%%%%%%%%%%%%%%%%%%%%%%%%%%%%%%%%%%%%%%%%%%%
% SD-3.5-Large -- State
%%%%%%%%%%%%%%%%%%%%%%%%%%%%%%%%%%%%%%%%%%%%%%%%%%%%%%%%%%%%%%%%%%%%%%%%%%%%%%%%%%%%%%%%%%
\begin{table}[ht]
\centering
\scriptsize
\setlength{\tabcolsep}{4pt}
\renewcommand{\arraystretch}{0.95}
\caption{Cluster membership for state prompts in SD-3.5-Large.}
\label{tab:cluster_composition_sd_35_large_state}
\begin{tabular}{r r p{0.68\linewidth}}
\toprule
\textbf{Cluster} & \textbf{Size} & \textbf{Locations} \\
\midrule
1 & 5 & Alaska, Colorado, Idaho, Montana, Oregon \\
2 & 2 & Florida, Hawaii \\
3 & 4 & North Dakota, South Dakota, Utah, Wyoming \\
4 & 4 & Arizona, California, Nevada, New Mexico \\
5 & 9 & Connecticut, Delaware, Maine, Maryland, Massachusetts, New Hampshire, Rhode Island, Vermont, Virginia \\
6 & 12 & Georgia, Illinois, Indiana, Michigan, Minnesota, Missouri, New Jersey, New York, Ohio, Pennsylvania, State of Washington, Wisconsin \\
7 & 4 & Arkansas, Kentucky, Tennessee, West Virginia \\
8 & 5 & Iowa, Kansas, Nebraska, Oklahoma, Texas \\
9 & 5 & Alabama, Louisiana, Mississippi, North Carolina, South Carolina \\
\bottomrule
\end{tabular}
\end{table}

%%%%%%%%%%%%%%%%%%%%%%%%%%%%%%%%%%%%%%%%%%%%%%%%%%%%%%%%%%%%%%%%%%%%%%%%%%%%%%%%%%%%%%%%%%
% SD-3.5-Large -- Capital
%%%%%%%%%%%%%%%%%%%%%%%%%%%%%%%%%%%%%%%%%%%%%%%%%%%%%%%%%%%%%%%%%%%%%%%%%%%%%%%%%%%%%%%%%%
\begin{table}[ht]
\centering
\scriptsize
\setlength{\tabcolsep}{4pt}
\renewcommand{\arraystretch}{0.95}
\caption{Cluster membership for capital prompts in SD-3.5-Large.}
\label{tab:cluster_composition_sd_35_large_capital}
\begin{tabular}{r r p{0.68\linewidth}}
\toprule
\textbf{Cluster} & \textbf{Size} & \textbf{Locations} \\
\midrule
1 & 4 & Atlanta, Austin, Indianapolis, Oklahoma City \\
2 & 10 & Baton Rouge, Boston, Des Moines, Jackson, Little Rock, Nashville, Raleigh, Sacramento, Springfield, Topeka \\
3 & 6 & Boise, Carson City, Cheyenne, Helena, Juneau, Santa Fe \\
4 & 4 & Denver, Honolulu, Phoenix, Salt Lake City \\
5 & 2 & Bismarck, Pierre \\
6 & 1 & Dover \\
7 & 2 & Annapolis, Charleston \\
8 & 8 & Augusta, Concord, Frankfort, Montgomery, Richmond, Salem, Tallahassee, Trenton \\
9 & 7 & Albany, Columbia, Columbus, Harrisburg, Hartford, Lansing, Providence \\
10 & 5 & Jefferson City, Lincoln, Madison, Montpelier, Saint Paul \\
11 & 1 & Olympia \\
\bottomrule
\end{tabular}
\end{table}

\clearpage
\section{Additional rankings by distance to the generic \texttt{USA} prompt}
\label{app:usa_distance_rankings}

Tables~\ref{tab:appendix_usa_distance_flux1_schnell_state}--\ref{tab:appendix_usa_distance_sd_35_large_capital}
report the full rankings of state and capital prompts by their FID to the generic
\texttt{USA} prompt. Lower values indicate greater similarity to the model-specific national prototype.

%%%%%%%%%%%%%%%%%%%%%%%%%%%%%%%%%%%%%%%%%%%%%%%%%%%%%%%%%%%%%%%%%%%%%%%%%%%%%%%%%%%%%%%%%%
% FLUX 1-\textit{schnell} -- State
%%%%%%%%%%%%%%%%%%%%%%%%%%%%%%%%%%%%%%%%%%%%%%%%%%%%%%%%%%%%%%%%%%%%%%%%%%%%%%%%%%%%%%%%%%
\begin{table}[ht]
\centering
\scriptsize
\setlength{\tabcolsep}{5pt}
\renewcommand{\arraystretch}{0.95}
\caption{State prompts ranked by FID to \texttt{USA} for FLUX 1-\textit{schnell}.}
\label{tab:appendix_usa_distance_flux1_schnell_state}
\begin{tabular}{r l r @{\qquad} r l r}
\toprule
\textbf{Rank} & \textbf{Location} & \textbf{Distance} &
\textbf{Rank} & \textbf{Location} & \textbf{Distance} \\
\midrule
1 & New Jersey & 331.8 & 26 & Tennessee & 1144.0 \\
2 & Michigan & 404.2 & 27 & Arkansas & 1151.2 \\
3 & Illinois & 414.1 & 28 & South Carolina & 1198.9 \\
4 & Minnesota & 545.8 & 29 & South Dakota & 1203.3 \\
5 & Texas & 599.4 & 30 & Louisiana & 1221.9 \\
6 & Indiana & 624.6 & 31 & Pennsylvania & 1296.2 \\
7 & State of New York & 636.2 & 32 & North Dakota & 1392.0 \\
8 & Oklahoma & 638.8 & 33 & Oregon & 1504.6 \\
9 & Ohio & 675.0 & 34 & Colorado & 1542.9 \\
10 & Connecticut & 675.9 & 35 & West Virginia & 1565.4 \\
11 & Massachusetts & 689.2 & 36 & New Hampshire & 1578.1 \\
12 & Missouri & 693.1 & 37 & Wyoming & 1580.9 \\
13 & Mississippi & 696.3 & 38 & Vermont & 1654.0 \\
14 & Kansas & 696.8 & 39 & Maine & 1655.0 \\
15 & Georgia & 703.9 & 40 & State of Washington & 1681.3 \\
16 & Wisconsin & 748.6 & 41 & New Mexico & 1700.2 \\
17 & Iowa & 812.0 & 42 & Idaho & 1729.3 \\
18 & Alabama & 847.0 & 43 & Montana & 1778.9 \\
19 & Maryland & 847.9 & 44 & Nevada & 1803.1 \\
20 & Nebraska & 867.8 & 45 & Utah & 1813.2 \\
21 & Virginia & 996.7 & 46 & California & 1961.9 \\
22 & Kentucky & 1026.3 & 47 & Arizona & 1970.6 \\
23 & North Carolina & 1075.9 & 48 & Florida & 1988.7 \\
24 & Rhode Island & 1096.9 & 49 & Alaska & 2143.4 \\
25 & Delaware & 1112.4 & 50 & Hawaii & 2594.5 \\
\bottomrule
\end{tabular}
\end{table}

%%%%%%%%%%%%%%%%%%%%%%%%%%%%%%%%%%%%%%%%%%%%%%%%%%%%%%%%%%%%%%%%%%%%%%%%%%%%%%%%%%%%%%%%%%
% FLUX 1-\textit{schnell} -- Capital
%%%%%%%%%%%%%%%%%%%%%%%%%%%%%%%%%%%%%%%%%%%%%%%%%%%%%%%%%%%%%%%%%%%%%%%%%%%%%%%%%%%%%%%%%%
\begin{table}[p]
\centering
\scriptsize
\setlength{\tabcolsep}{5pt}
\renewcommand{\arraystretch}{0.95}
\caption{Capital prompts ranked by FID to \texttt{USA} for FLUX 1-\textit{schnell}.}
\label{tab:appendix_usa_distance_flux1_schnell_capital}
\begin{tabular}{r l r @{\qquad} r l r}
\toprule
\textbf{Rank} & \textbf{Location} & \textbf{Distance} &
\textbf{Rank} & \textbf{Location} & \textbf{Distance} \\
\midrule
1 & Madison & 408.5 & 26 & Indianapolis & 1051.3 \\
2 & Raleigh & 467.5 & 27 & Lincoln & 1090.7 \\
3 & Jackson & 484.7 & 28 & Providence & 1178.1 \\
4 & Nashville & 501.1 & 29 & Salem & 1209.0 \\
5 & Little Rock & 539.5 & 30 & Charleston & 1234.0 \\
6 & Boston & 557.6 & 31 & Augusta & 1235.3 \\
7 & Austin & 567.5 & 32 & Columbus & 1238.3 \\
8 & Montgomery & 602.9 & 33 & Annapolis & 1244.5 \\
9 & Hartford & 615.5 & 34 & Cheyenne & 1305.0 \\
10 & Baton Rouge & 631.8 & 35 & Carson City & 1350.5 \\
11 & Albany & 635.9 & 36 & Richmond & 1410.8 \\
12 & Des Moines & 638.4 & 37 & Bismarck & 1430.5 \\
13 & Trenton & 643.2 & 38 & Olympia & 1446.2 \\
14 & Columbia & 644.1 & 39 & Phoenix & 1543.3 \\
15 & Lansing & 707.6 & 40 & Jefferson City & 1564.4 \\
16 & Oklahoma City & 744.4 & 41 & Juneau & 1653.5 \\
17 & Salt Lake City & 768.1 & 42 & Santa Fe & 1702.0 \\
18 & Denver & 832.4 & 43 & Concord & 1741.2 \\
19 & Atlanta & 887.2 & 44 & Honolulu & 1757.2 \\
20 & Sacramento & 914.3 & 45 & Helena & 2040.4 \\
21 & Topeka & 953.8 & 46 & Montpelier & 2481.1 \\
22 & Harrisburg & 981.6 & 47 & Pierre & 2522.3 \\
23 & Tallahassee & 989.6 & 48 & Dover & 2530.8 \\
24 & Springfield & 993.8 & 49 & Saint Paul & 2730.7 \\
25 & Boise & 1036.7 & 50 & Frankfort & 2968.3 \\
\bottomrule
\end{tabular}
\end{table}

%%%%%%%%%%%%%%%%%%%%%%%%%%%%%%%%%%%%%%%%%%%%%%%%%%%%%%%%%%%%%%%%%%%%%%%%%%%%%%%%%%%%%%%%%%
% Stable Diffusion 3.5-L -- State
%%%%%%%%%%%%%%%%%%%%%%%%%%%%%%%%%%%%%%%%%%%%%%%%%%%%%%%%%%%%%%%%%%%%%%%%%%%%%%%%%%%%%%%%%%
\begin{table}[p]
\centering
\scriptsize
\setlength{\tabcolsep}{5pt}
\renewcommand{\arraystretch}{0.95}
\caption{State prompts ranked by FID to \texttt{USA} for Stable Diffusion 3.5-L.}
\label{tab:appendix_usa_distance_sd_35_large_state}
\begin{tabular}{r l r @{\qquad} r l r}
\toprule
\textbf{Rank} & \textbf{Location} & \textbf{Distance} &
\textbf{Rank} & \textbf{Location} & \textbf{Distance} \\
\midrule
1 & Illinois & 413.7 & 26 & South Carolina & 1875.6 \\
2 & Minnesota & 796.2 & 27 & Delaware & 1908.9 \\
3 & New York & 829.2 & 28 & Arkansas & 1919.1 \\
4 & Georgia & 843.5 & 29 & Tennessee & 2025.9 \\
5 & Indiana & 848.3 & 30 & Rhode Island & 2120.5 \\
6 & Ohio & 1001.6 & 31 & New Hampshire & 2194.9 \\
7 & New Jersey & 1006.4 & 32 & West Virginia & 2236.9 \\
8 & Michigan & 1012.5 & 33 & California & 2282.8 \\
9 & Oklahoma & 1109.8 & 34 & Vermont & 2298.8 \\
10 & Missouri & 1123.1 & 35 & Kentucky & 2333.2 \\
11 & Wisconsin & 1180.1 & 36 & New Mexico & 2342.1 \\
12 & Texas & 1245.5 & 37 & Maine & 2498.2 \\
13 & Alabama & 1370.9 & 38 & Florida & 2546.2 \\
14 & Pennsylvania & 1444.7 & 39 & Colorado & 2610.5 \\
15 & Nebraska & 1516.0 & 40 & Utah & 2713.2 \\
16 & Massachusetts & 1584.6 & 41 & South Dakota & 2903.0 \\
17 & Virginia & 1591.6 & 42 & Alaska & 2993.6 \\
18 & Kansas & 1620.2 & 43 & Oregon & 3014.9 \\
19 & North Carolina & 1628.6 & 44 & Idaho & 3052.2 \\
20 & Louisiana & 1634.9 & 45 & Montana & 3083.1 \\
21 & State of Washington & 1790.5 & 46 & Wyoming & 3088.3 \\
22 & Mississippi & 1808.0 & 47 & Nevada & 3131.7 \\
23 & Connecticut & 1809.6 & 48 & Hawaii & 3139.8 \\
24 & Maryland & 1813.9 & 49 & North Dakota & 3207.3 \\
25 & Iowa & 1841.0 & 50 & Arizona & 3352.9 \\
\bottomrule
\end{tabular}
\end{table}

%%%%%%%%%%%%%%%%%%%%%%%%%%%%%%%%%%%%%%%%%%%%%%%%%%%%%%%%%%%%%%%%%%%%%%%%%%%%%%%%%%%%%%%%%%
% Stable Diffusion 3.5-L -- Capital
%%%%%%%%%%%%%%%%%%%%%%%%%%%%%%%%%%%%%%%%%%%%%%%%%%%%%%%%%%%%%%%%%%%%%%%%%%%%%%%%%%%%%%%%%%
\begin{table}[p]
\centering
\scriptsize
\setlength{\tabcolsep}{5pt}
\renewcommand{\arraystretch}{0.95}
\caption{Capital prompts ranked by FID to \texttt{USA} for Stable Diffusion 3.5-L.}
\label{tab:appendix_usa_distance_sd_35_large_capital}
\begin{tabular}{r l r @{\qquad} r l r}
\toprule
\textbf{Rank} & \textbf{Location} & \textbf{Distance} &
\textbf{Rank} & \textbf{Location} & \textbf{Distance} \\
\midrule
1 & Indianapolis & 570.9 & 26 & Lincoln & 1502.2 \\
2 & Austin & 645.2 & 27 & Richmond & 1552.4 \\
3 & Boston & 662.8 & 28 & Honolulu & 1559.3 \\
4 & Raleigh & 665.9 & 29 & Trenton & 1594.8 \\
5 & Sacramento & 705.3 & 30 & Frankfort & 1654.7 \\
6 & Columbia & 762.8 & 31 & Augusta & 1661.3 \\
7 & Little Rock & 771.0 & 32 & Boise & 1711.5 \\
8 & Jackson & 775.0 & 33 & Phoenix & 1797.0 \\
9 & Oklahoma City & 775.3 & 34 & Concord & 1936.5 \\
10 & Des Moines & 788.8 & 35 & Helena & 1944.6 \\
11 & Baton Rouge & 851.5 & 36 & Salem & 2042.5 \\
12 & Atlanta & 903.2 & 37 & Carson City & 2069.6 \\
13 & Nashville & 944.7 & 38 & Montpelier & 2091.8 \\
14 & Hartford & 1007.6 & 39 & Madison & 2105.1 \\
15 & Topeka & 1028.8 & 40 & Cheyenne & 2133.5 \\
16 & Columbus & 1100.5 & 41 & Jefferson City & 2161.0 \\
17 & Denver & 1101.0 & 42 & Santa Fe & 2283.7 \\
18 & Lansing & 1105.5 & 43 & Annapolis & 2394.3 \\
19 & Albany & 1139.4 & 44 & Charleston & 2524.3 \\
20 & Providence & 1150.9 & 45 & Saint Paul & 2555.7 \\
21 & Springfield & 1242.7 & 46 & Juneau & 2607.0 \\
22 & Salt Lake City & 1388.5 & 47 & Pierre & 2962.8 \\
23 & Tallahassee & 1389.7 & 48 & Bismarck & 3192.8 \\
24 & Harrisburg & 1422.7 & 49 & Olympia & 3430.7 \\
25 & Montgomery & 1490.9 & 50 & Dover & 3839.6 \\
\bottomrule
\end{tabular}
\end{table}

\end{document}